\pdfoutput=1
\documentclass[11pt]{article}

\usepackage[preprint]{acl}

\usepackage{times}
\usepackage{latexsym}
\usepackage[T1]{fontenc}
\usepackage[utf8]{inputenc}
\usepackage{microtype}
\usepackage{mathptmx}

\usepackage{graphicx}
\usepackage{xurl}
\usepackage{booktabs}
\usepackage{amsmath}
\usepackage{amssymb}
\usepackage{amsthm}
\usepackage{mathtools}
\usepackage{bm}
\usepackage{array}
\usepackage{enumitem}
\usepackage{algorithm}
\usepackage{algpseudocode}
\usepackage{wrapfig}
\usepackage{xcolor}
\usepackage{colortbl}
\usepackage{siunitx}
\usepackage{makecell}
\usepackage[capitalize,nameinlink,noabbrev]{cleveref}

\graphicspath{{./}}

\definecolor{best}{RGB}{217,234,247}
\definecolor{rowtint}{RGB}{246,248,250}
\newcommand{\rowbest}{\rowcolor{best}}
\newcommand{\rowtint}{\rowcolor{rowtint}}
\newcommand{\cellbest}[1]{\cellcolor{best}\bfseries #1}

\newcommand{\op}[1]{\emph{#1}}

\newcolumntype{N}{S[table-format=1.2]}

\renewcommand{\arraystretch}{1.15}

\newcommand{\term}[1]{\textsc{#1}}
\newcommand{\STATE}{\State}
\newcommand{\FORALL}{\ForAll}
\newcommand{\IF}{\If}
\newcommand{\ELSIF}{\ElsIf}
\newcommand{\ENDIF}{\EndIf}
\newcommand{\ENDFOR}{\EndFor}

\title{LatticeMind: A Conflict-Aware Memory Primitive for Multi-Agent Systems}

\author{
  Heng Zhou\textsuperscript{1}\thanks{\hspace{0.2em}Equal contribution.} \quad
  Lian Zhang\textsuperscript{2}\footnotemark[1] \quad
  Yutao Fan\textsuperscript{3}\footnotemark[1] \quad
  Tiancheng He\textsuperscript{4} \\
  \bfseries
  Siki Chen\textsuperscript{5} \quad
  Hejia Geng\textsuperscript{5} \quad
  Philip Torr\textsuperscript{5} \quad
  Zhenfei Yin\textsuperscript{5} \\[0.4em]
  \normalsize
  \textsuperscript{1}University of Science and Technology of China \quad
  \textsuperscript{2}University of Illinois Urbana-Champaign \\
  \normalsize
  \textsuperscript{3}Shanghai AI Laboratory \quad
  \textsuperscript{4}Beijing University of Posts and Telecommunications \quad
  \textsuperscript{5}University of Oxford
}

\begin{document}
\maketitle

\begin{abstract}
Multi-agent LLM systems often fail not for lack of candidate answers, but because they have no persistent mechanism for deciding which incompatible claim should currently be trusted. Majority vote, debate, and judge-based selection choose an output without recording which claim wins, which is contested, or why a later update supersedes it. We present \term{LatticeMind}, a conflict-aware structured memory that handles contradiction at write time. It maintains explicit item status, applies cheap symbolic conflict checks, and invokes LLM reconciliation only for unresolved semantic cases. On a label-blind ConflictBank evaluation that removes source-name hints, LatticeMind reaches 0.97 accuracy versus 0.61 for the strongest aggregation baseline, with the gap significant at $p<10^{-6}$ by paired McNemar test. Ablations show that removing the checker or the reconciler costs 12 to 14 points. On four secondary planning benchmarks the picture is mixed: LatticeMind beats naive merge on three of four, but does not replace deliberation methods on tasks rewarding iterative search.

\end{abstract}

\section{Introduction}\label{sec:introduction}

Consider a multi-agent system answering a question about an updated approval policy. Agent A reads the authoritative policy document and writes the new version. Agent B has only seen an outdated note and asserts the legacy policy. Agent C cites a recent third-party report whose recency is misleading. Standard aggregation strategies, including majority vote~\citep{wang2023selfconsistency}, debate~\citep{du2024debate}, and judge-based selection~\citep{zheng2023judging}, decide what to output, but they leave no trace of which claim won, which one was contested, or why. The next time a similar question arrives, the same conflict has to be relitigated from scratch. This is the failure mode we target: not a shortage of context, but the absence of a persistent mechanism for managing disagreement over time.

We argue that the missing primitive is conflict-aware memory. Existing agent-memory architectures~\citep{packer2023memgpt,sumers2024coala,park2023generative} cover retrieval, compression, and reflection, but none treats contradiction handling as a first-class memory update operation. Free-form summarization can produce coherent prose while erasing why a fact is trusted or which older claim it replaced; raw concatenation keeps everything but provides no suppression mechanism. A recent survey organizes agent memory by form, function, and temporal dynamics~\citep{hu2025memoryagents}, yet none of its axes treats \emph{conflict status} as a first-class property: which claim currently wins, which is contested, and which is superseded. Combined with the rise of knowledge conflicts as a core LLM challenge~\citep{xu2024knowledge}, this motivates an explicit conflict-handling primitive. The same problem reappears in long-horizon software projects, where repositories evolve and different scopes such as production, staging, and feature branches temporarily disagree about the current state.

We therefore study a narrow question. Should contradiction handling in multi-agent systems be treated as a memory problem rather than an answer-selection problem? Our answer is \term{LatticeMind}, a structured memory framework with explicit item status, symbolic conflict checks, and selective LLM reconciliation. Cheap symbolic checks detect mechanical conflicts such as dependency cycles or resource violations. The reconciler is invoked only when needed, and it classifies each remaining conflict as a credibility decision, where one claim should defeat another, or as a coordination decision, where multiple locally valid proposals must coexist while planning is revised. Our empirical evaluation focuses on the credibility regime; coordination handling is present as a conservative safety override rather than as a validated claim.

We structure the paper around one main claim and one application study. On ConflictBank~\citep{su2024conflictbank}, evaluated under a label-blind protocol that anonymizes source names to Source A/B/C and removes any trust hint, LatticeMind reaches 0.97 accuracy versus 0.63 for single-agent inference and 0.61 for the strongest aggregation baseline, with every gap significant at $p<10^{-6}$ by paired McNemar test. Secondary planning results are deliberately mixed, mapping where conflict-aware memory helps. The application study then shows the same update rules transfer to long-horizon software-state tracking, with a learned-operator study localizing the remaining gap to extraction rather than to conflict resolution.

Our contributions are fourfold:
\begin{itemize}[label=$\diamond$, leftmargin=1.4em, itemsep=2pt, topsep=3pt, parsep=0pt]
    \item We identify contradiction handling as a missing primitive in multi-agent systems and formulate it as a structured memory update problem rather than an answer-aggregation problem.
    \item We introduce \term{LatticeMind}, which combines explicit status tracking, symbolic conflict checks, and selective LLM reconciliation in a single update loop.
    \item We report strong ConflictBank gains with statistical tests and a single-pass ablation that isolates the complementary roles of the checker and the reconciler.
    \item We instantiate the same rules for software-state tracking and a learned-operator study that localizes the remaining gap to extraction rather than to conflict resolution.
\end{itemize}

\section{Related Work}\label{sec:related_work}

\subsection{Memory Systems and Benchmarks}
\label{subsec:memory}

Memory has become a central concern for LLM-based agents. The recent survey by \citet{hu2025memoryagents} organizes the field along three axes: memory form (token-level, parametric, latent), function (factual, experiential, working), and temporal dynamics. It explicitly separates agent memory from LLM memory, retrieval-augmented generation~\citep{lewis2020retrieval}, and context engineering. None of its functional categories, however, represents or resolves contradictory claims over time, which is the gap LatticeMind targets. Earlier architectures explore different points in this space: MemGPT~\citep{packer2023memgpt} treats the LLM as an OS with explicit paging; Generative Agents~\citep{park2023generative} maintain a memory stream of observations and reflections; Reflexion~\citep{shinn2023reflexion} uses verbal self-reflection; and CoALA~\citep{sumers2024coala} unifies working, episodic, semantic, and procedural stores. More recent token-level systems extend this store: Mem0~\citep{chhikara2025mem0} consolidates salient facts, A-MEM~\citep{xu2025amem} self-organizes notes into a linked graph, and HippoRAG~\citep{gutierrez2024hipporag} indexes memory through a knowledge graph, while a parametric line edits facts directly into model weights~\citep{meng2022rome}. Three recent systems are closest to ours. Zep~\citep{rasmussen2025zep} stores memory as a temporal knowledge graph but exposes no explicit per-item status (proposed, contested, superseded); supersession is implicit in the temporal index. MIRIX~\citep{wang2025mirix} composes multi-agent memory at the retrieval-and-summarization layer rather than at a shared write target with conflict resolution. FaithfulRAG~\citep{zhang2025faithfulrag} resolves fact-level conflicts just before generation rather than persisting the resolution for later agents. LatticeMind instead makes conflict resolution a write-time operation on shared memory with explicit item status, so the next reader sees a resolved state rather than relitigating the conflict.

Recent memory benchmarks improve evaluation but mostly target conversational or single-agent settings. LoCoMo~\citep{maharana2024locomo} and LongMemEval~\citep{wu2025longmemeval} cover chat memory, MemBench~\citep{tan2025membench} covers factual and reflective memory, MemoryAgentBench~\citep{hu2025memoryagentbench} measures incremental multi-turn memory in agents, and ConflictBank~\citep{su2024conflictbank} targets knowledge conflicts. The companion survey of \citet{xu2024knowledge} taxonomizes context-memory, inter-context, and intra-memory conflicts. A parallel line builds benchmarks over knowledge that changes underneath the model, so that yesterday's correct answer becomes today's stale one~\citep{zhou2025livesearchbench}; this shares our concern with supersession but measures retrieval against a moving corpus rather than the memory update itself. None of these benchmarks isolates the multi-agent setting where memory must decide which of several incompatible facts to trust, under what scope, and with what provenance.

\subsection{Multi-Agent and Long-Horizon Software Settings}
\label{subsec:multiagent_software}

Multi-agent LLM frameworks built on ReAct~\citep{yao2023react} have grown rapidly. AutoGen~\citep{wu2023autogen}, CAMEL~\citep{li2023camel}, MetaGPT~\citep{hong2024metagpt}, and ChatDev~\citep{qian2024chatdev} focus on task decomposition, role assignment, and communication protocols, leaving disagreement handling to downstream aggregation or human oversight~\citep{guo2024multiagent,xi2023rise}. Later work makes the organization itself adaptive, either by letting a reward signal reshape the agent topology~\citep{zhou2025reso} or by training the policy that drives the agents~\citep{zhang2025landscape}; both optimize how work is assigned rather than how conflicting results are reconciled once they are written. Long-horizon software benchmarks such as SWE-bench~\citep{jimenez2024swebench} and LongCLI-Bench~\citep{feng2026longclibench} measure end-to-end implementation under realistic execution constraints, and recent agent systems such as SWE-agent~\citep{yang2024sweagent}, Agentless~\citep{xia2024agentless}, and OpenHands~\citep{wang2024openhands} target patch generation directly. Our goal is complementary. Rather than benchmarking patch generation or bug fixing, we isolate the memory subproblem that precedes those tasks, maintaining a consistent and queryable view of evolving project state over long trajectories, with software-specific slots, provenance-aware queries, scope-sensitive retrieval, and budgeted downstream artifacts. We therefore do not compare against SWE-agent, Agentless, or OpenHands directly, since their target task is end-to-end issue resolution rather than auditable state maintenance.

\subsection{Aggregation Versus Persistent Memory}
\label{subsec:aggregation_vs_memory}

Many LLM systems handle multiple sources via answer-level aggregation: majority vote~\citep{wang2023selfconsistency}, multi-agent debate~\citep{du2024debate}, or judge-based selection~\citep{zheng2023judging}. We compare against these directly. They operate over answers rather than an auditable persistent memory with explicit item status, scope, and supersession. The analogy to conflict-free replicated data types~\citep{shapiro2011conflict} is intentional: as CRDTs maintain eventual consistency through algebraic merge rules, LatticeMind maintains a consistent memory view through explicit status transitions and conflict-aware update policies.

\section{Method}\label{sec:method}

\subsection{Problem Formulation}
\label{subsec:problem}

We consider long-horizon multi-agent settings in which a system observes a sequence of information updates \(\mathcal{S} = (S_1, \dots, S_T)\).
An update may come from another agent, a note, or a tool result. The core difficulty is that updates may contradict each other, differ in authority, or imply incompatible actions. The goal is therefore to maintain a memory \(M_T\) that remains auditable and queryable despite contradiction, supersession, and coordination failure.

We model each memory item as \(m = (k, c, \rho, \tau, \eta)\), where \(k\) is a key or routing handle, \(c\) is the content, \(\rho\) is evidence metadata, \(\tau\) is a timestamp, and \(\eta\) is a status drawn from \{\textsc{Proposed}, \textsc{Confirmed}, \textsc{Contested}, \textsc{Superseded}\}.
The memory problem is to update these items so that currently trusted facts become easy to retrieve, weaker or incompatible claims are not silently lost, and important transitions remain inspectable.

\paragraph{Running example.}
Consider three agents updating a project's approval policy. Agent A reads the authoritative policy document and writes that approvals should follow a new path. Agent B trusts an outdated note and writes that the legacy path is still active. Agent C proposes a Tuesday rollout that overlaps an already accepted deployment window. A summary-only system would flatten these into a single answer immediately. LatticeMind instead stores each finding as a separate item, routes the two policy claims into a credibility comparison, flags the rollout overlap through the symbolic checker, and preserves the losing or blocked items as contested rather than deleting them.

\begin{figure*}[t]
    \centering
    \includegraphics[width=0.95\textwidth]{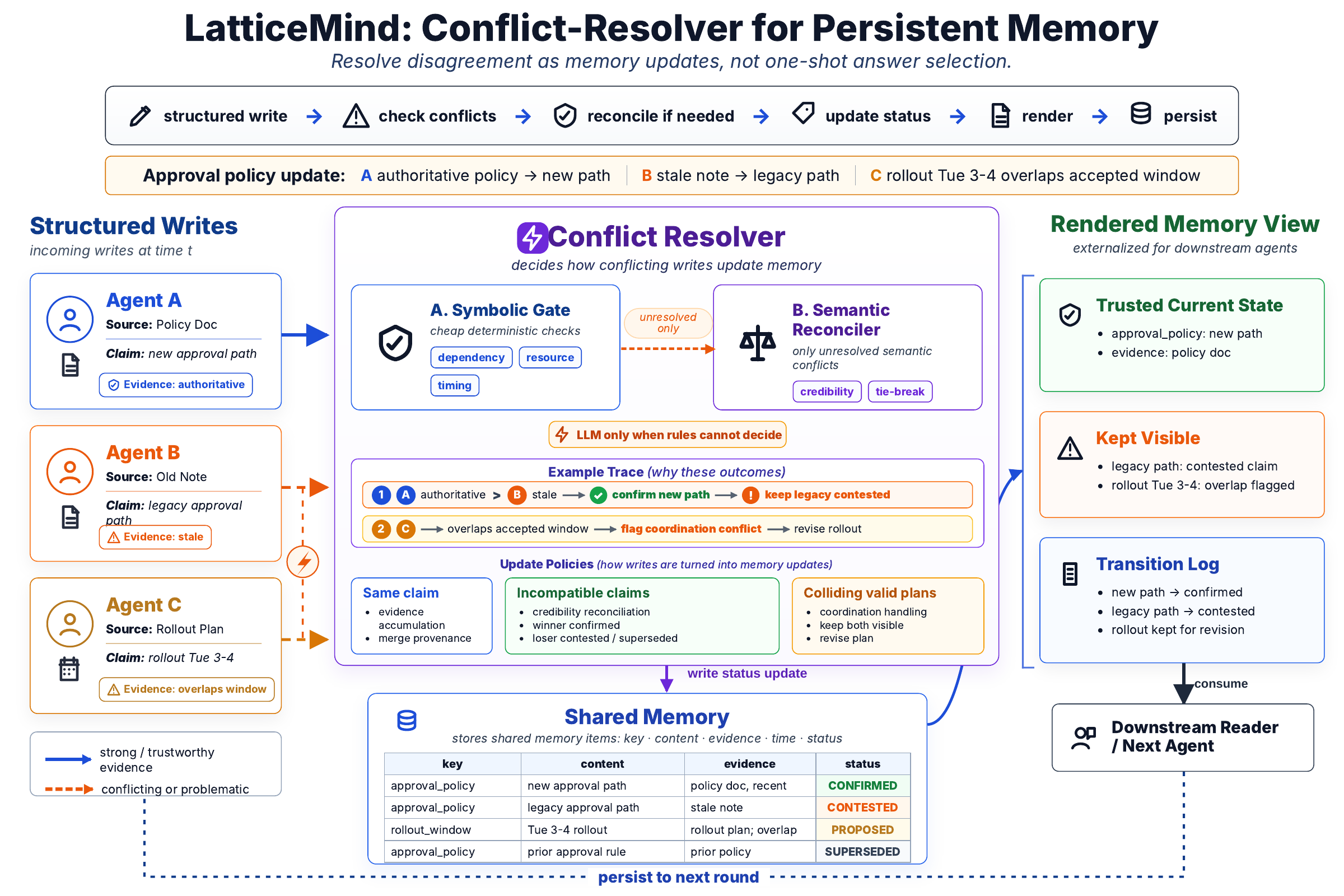}
    \caption{LatticeMind overview on an approval-policy example. Agents write structured findings into shared memory; a symbolic checker catches mechanical conflicts and selective LLM reconciliation resolves the rest, separating credibility from coordination; the system then renders an auditable memory view.}
    \label{fig:latticemind_overview}
\end{figure*}

\subsection{Proposed Approach}
\label{subsec:approach}

LatticeMind has three ingredients: structured writes, conflict-aware update rules, and rendered views for downstream use. \Cref{fig:latticemind_overview} sketches the loop, and \cref{alg:update_loop} states it precisely.

\paragraph{Structured writes.}
Agents produce findings such as \texttt{FACT}, \texttt{CONSTRAINT}, and \texttt{SUB\_PLAN}; these are written as explicit memory entries rather than immediately flattened into a final answer.
For the software-state application study, we additionally extract normalized state claims of the form \(z = (u, \sigma, v, \beta, \epsilon, \rho, \tau)\), where \(u\) is an entity, \(\sigma\) is a slot, \(v\) is the slot value, \(\beta\) is branch scope, \(\epsilon\) is environment, \(\rho\) is evidence metadata, and \(\tau\) is timestamp. The canonical key for a state claim is
\begin{equation}
    k(z) = (u, \sigma, \beta, \epsilon).
    \label{eq:key}
\end{equation}
\vspace{-0.5em}
This keying lets the application-specific memory reason over one scope-local slot at a time instead of treating the entire history as undifferentiated text.

\paragraph{Conflict-aware update rules.}
LatticeMind uses different update rules for different conflict types. A symbolic checker first catches mechanical violations such as dependency cycles or resource collisions. The reconciler then classifies the remaining semantic conflicts, in the spirit of routing an input to the procedure that suits it rather than applying one fixed strategy~\citep{zhou2026select}. Credibility conflicts, where one claim should beat another and which are the focus of our empirical evaluation, trigger evidence-weighted supersession. Coordination conflicts, where multiple locally valid proposals collide over a shared resource, trigger a conservative safety override that keeps both candidates visible and re-runs planning rather than deleting either. The coordination branch is included as a design-level safety mechanism, and we do not present it as an empirically validated improvement.

For the software-state application study, incoming claims are first routed by the canonical key in \cref{eq:key}. If a new value matches the current value after normalization, we merge evidence and provenance. If it differs, we compare evidence strength before considering semantic similarity. Our score is \(s(z) = w(\op{evidence-type}) + 40 \cdot \mathbf{1}[\texttt{git\_commit} \neq \emptyset]\), where \(w(\cdot)\) gives higher weight to code-backed facts than to human notes or explicitly stale observations. In the implementation, \op{code-change} and \op{incident-hotfix} are strongest, followed by configuration and runtime observations; \op{stale-observation} is weakest.

If an incoming claim is stronger than the existing claim for the same key, it supersedes the old state and becomes the new current value. If it is weaker, it is retained as a contested observation rather than silently discarded. When evidence strength ties, we use recency; only then do we fall back to an LLM judge.

LatticeMind defines four update policies, each triggered by a distinct conflict pattern. Credibility reconciliation fires when incompatible claims describe the same fact or answer; the stronger claim is confirmed and the weaker one is preserved as \texttt{SUPERSEDED} or \texttt{CONTESTED} with full provenance. Coordination handling fires on dependency cycles or resource overlap among otherwise useful proposals; candidate items remain visible, a conflict signal is attached, and planning is rerun rather than deleting proposals prematurely. Slot-first state merge fires when two claims share the same \((\text{entity}, \text{slot}, \text{branch}, \text{env})\) key but disagree on value; the winner becomes current state and the loser is marked contested. Evidence accumulation fires when the same key and normalized value reappear; sources, timestamps, and supporting evidence are merged into the existing entry rather than spawning duplicates.

\paragraph{Rendered memory.}
The final memory view is externalized rather than kept as an inaccessible hidden structure. For multi-agent reasoning, the rendered view contains the resolved findings that survive conflict checking and reconciliation. For software-state tasks, the renderer produces current scope state, contested observations, and transitions; under tight budgets it groups tightly coupled slots such as database with ORM or deployment with compute/hosting. In both cases, downstream readers only see the rendered memory document, so the evaluation measures the usefulness of the memory representation itself.

\paragraph{Learned operators for software-state memory.}
To probe whether the memory update rules themselves are learnable rather than purely hand-written, we also train lightweight LoRA~\citep{hu2022lora} adapters for three software-state operators:
\op{extract-state},
\op{compare-claims}, and
\op{update-memory}.
\op{extract-state} maps raw notes into normalized slot claims;
\op{compare-claims} predicts a relation label and winner for two competing claims; and
\op{update-memory} predicts whether a new claim should be added, merged, supersede the current state, or remain contested.
These learned operators are used only in the repository replay study, not in the main ConflictBank result. This separation lets us ask a sharper question: which parts of the software-state pipeline transfer across repositories once the basic conflict-aware abstraction is fixed?

\begin{algorithm}[t]
\caption{LatticeMind update loop}
\label{alg:update_loop}
\begin{algorithmic}[1]
\STATE \textbf{Input:} current memory \(M\), new update \(S_t\)
\STATE Extract structured items \(E_t\) from \(S_t\)
\FORALL{item \(e \in E_t\)}
    \IF{\(e\) is an agent finding}
        \STATE append \(e\) to \(M\) with explicit status and evidence
    \ELSIF{\(e\) is a software-state claim}
        \STATE route \(e\) by canonical key \(k(e)\)
        \STATE compare against current value using evidence score, recency, and LLM tie-break only if needed
        \STATE mark the winner current and keep the loser as contested or superseded
    \ENDIF
\ENDFOR
\STATE Run symbolic checks for dependency or resource conflicts
\IF{violations exist}
    \STATE classify them as credibility or coordination conflicts
    \STATE invoke LLM reconciliation only for the unresolved semantic cases
\ENDIF
\STATE Render \(M\) into current state, contested items, and transition view
\STATE \textbf{Return:} rendered memory document
\end{algorithmic}
\end{algorithm}

\subsection{Why Structured Memory?}
\label{subsec:why_slot_first}

The core design choice behind LatticeMind is that many long-horizon failures come from implicit memory. If multiple findings are collapsed into a paragraph, or if all history is retained as raw text, the system has no explicit notion of which fact currently wins, which claim is contested, or which update caused the transition. Structured memory reduces this ambiguity in three ways. First, it forces incompatible items to meet at an explicit key or checker-defined conflict point. Second, it preserves supersession and contestation rather than deleting disagreement. Third, it makes later provenance and temporal reasoning possible because changes are stored instead of only narrated.

\section{Experiments}\label{sec:experiments}

\subsection{Experimental Setup}
\label{subsec:setup}

We organize the experiments around one main claim and one application study.

\paragraph{Main public contradiction benchmark.}
Our main public benchmark is ConflictBank~\citep{su2024conflictbank}, where multiple evidence sources disagree and the model must select the correct answer. We compare LatticeMind against five baselines: \textsc{Single-Agent} reads all evidence in one pass, \textsc{No-Merge} merges multiple agent outputs naively, and \textsc{Majority Vote}, \textsc{Debate}, and \textsc{Judge} aggregate at the answer level. All methods are evaluated on the same 75 examples with Qwen3-Max.

\paragraph{Label-blind protocol.}
The original ConflictBank construction names each source by its role (\emph{authoritative source}, \emph{alternative source}, \emph{recent report}), and these names are a potential oracle cue: a reader could prefer ``authoritative source'' regardless of memory state. To remove this confound, we evaluate every method under a label-blind protocol that renames the three sources to neutral identifiers \emph{Source A}, \emph{Source B}, and \emph{Source C}, drops any trust hint from the reader prompts, and uses the same neutral labels across LatticeMind and all baselines. All ConflictBank methods consume the same evidence pool under the same neutral identifiers. The principal difference at the answer-aggregation layer is whether incompatible findings are resolved through an explicit persistent memory update (LatticeMind) or aggregated at the answer level (all baselines); the appendix quantifies the additional checker and reconciler overhead and reports a fairness check via an evidence-tag ablation (\cref{app:label_blind_diff} and the discussion of same-pool baseline parity). All ConflictBank numbers reported in the main paper are produced under this protocol.

\paragraph{Secondary public benchmarks.}
We also report NaturalPlan Calendar~\citep{zheng2024naturalplan}, PlanBench~\citep{valmeekam2023planbench}, TravelPlanner~\citep{xie2024travelplanner}, and REALM JSSP~\citep{geng2025realmbench}. These are not contradiction benchmarks in the same sense as ConflictBank; we include them to map the boundary of the method.

\paragraph{Application study: software-state tracking.}
As an application of the same memory design, we instantiate LatticeMind as a slot-first state memory for long-horizon repository tracking. We compare four methods: \textsc{Concat} concatenates session notes, \textsc{LLM-Merge} produces a one-pass summary, \textsc{LLM-Merge-Inc} updates a summary incrementally, and \textsc{StateMemory} is the slot-first LatticeMind instantiation. The benchmark contains three scenario families: \op{authority-overwrite}, \op{scope-divergence}, and \op{incident-timeline}. We focus on the strongest stress test, \textsc{budgeted long-noise}, where scenarios are lengthened with stale, branch, and documentation noise; methods produce documents of at most 1700 characters and the reader sees at most 1400 characters.

\paragraph{Metrics.}
ConflictBank is measured by answer accuracy. Software-state query answering evaluates \op{current-state}, \op{scoped-state}, \op{provenance}, \op{temporal}, and \op{stale-suppression}. Answers are generated by an LLM reader but scored deterministically with required, forbidden, and any-of token checks.

\paragraph{Real-repository bridge.}
To partially improve external validity, we add a replay-based bridge over two real Codex commit trajectories: a permissions-flow rollout and an exec/app-server rollout. Scenarios are built from real commit subjects, summaries, dates, and touched files, then replayed through the same state-memory pipeline.

\paragraph{Learned repo-operator study.}
We train 4-bit QLoRA~\citep{dettmers2023qlora} adapters on \texttt{Qwen2.5-14B-Instruct}~\citep{qwen2024qwen25} for the three operator tasks of \cref{sec:method}: \op{extract-state}, \op{compare-claims}, and \op{update-memory}. We evaluate them in a five-way leave-one-out setting over replayed Codex trajectories, holding out one trajectory and reporting schema-only scores on the unseen trajectory.

\subsection{Main Public Results}
\label{subsec:results}

\begin{figure}[t]
    \centering
    \includegraphics[width=\columnwidth]{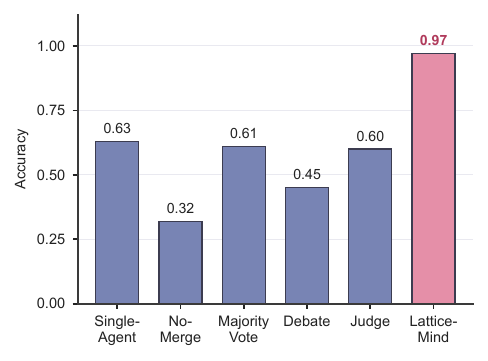}
    \caption{Label-blind ConflictBank accuracy across methods (75 examples). LatticeMind reaches 0.97, far above the strongest baseline overall (Single-Agent, 0.63) and the strongest answer-aggregation baseline (Majority Vote, 0.61). All baselines are paired-McNemar significantly below LatticeMind at $p<10^{-6}$.}
    \label{fig:conflictbank}
\end{figure}

\cref{fig:conflictbank} is the core result of the paper. Under the label-blind protocol described in \cref{subsec:setup}, in which the three evidence sources are presented to every method as Source A, Source B, and Source C without any role-revealing names or trust hints, LatticeMind reaches 0.97. The strongest answer-level baselines cluster more than 30 points below: Single-Agent at 0.63, Majority Vote at 0.61, and Judge at 0.60; \textsc{No-Merge} and \textsc{Debate} lag further at 0.32 and 0.45. The discordant-pair counts in \cref{tab:stats_main} are highly asymmetric: e.g.\ \textsc{No-Merge} is wrong on 49 examples that LatticeMind answers correctly and never wins on the cases LatticeMind misses. The advantage is therefore not produced by the prompt naming any source as authoritative; it comes from forcing incompatible findings through an explicit memory update and reconciliation step before rendering an answer.

\subsection{Ablation on Contradiction Handling}
\label{subsec:ablation}

\begin{table}[t]
    \centering
    \small
    \caption{ConflictBank single-pass ablation, 50 samples. Conflict rate is the fraction of examples where the symbolic checker flags an unresolved conflict in the rendered memory at answer time (lower is better). The last row is not an ablation: it re-runs the full configuration and serves as a run-to-run noise control (see text).}
    \label{tab:conflictbank_ablation}
    \begin{tabular}{lNN}
        \toprule
        Configuration & {Accuracy $\uparrow$} & {Conflict $\downarrow$} \\
        \midrule
        \textbf{Full}      & \cellbest{0.96} & \cellbest{0.00} \\
        \rowtint $-$checker     & 0.84 & 0.10 \\
        $-$reconciler           & 0.82 & 0.08 \\
        \rowtint Full (repeat)  & 0.96 & 0.02 \\
        \bottomrule
    \end{tabular}
\end{table}

\Cref{tab:conflictbank_ablation} reports the single-pass setting used by the final system. Point estimates drop 12 to 14 points when either the checker or the reconciler is removed. McNemar tests on the per-sample correctness array (\cref{app:stats}) confirm the reconciler ablation at $p<0.05$, while the checker ablation is directionally consistent at $p=0.07$ (limited by the 50-sample size). The last row deserves an explicit caveat. It was originally intended as an evidence-label ablation, stripping role names such as ``authoritative source'' from the evidence attached to each finding. Under the label-blind protocol those role names are already absent, so the configuration reduces to a second run of the full system. We report it as a run-to-run noise control: it lands on the same accuracy with zero discordant pairs, which bounds sampling variance but says nothing about the contribution of evidence labels. That question is answered instead by \cref{tab:label_blind_diff}, where removing role names from the reader prompt costs every baseline 8 to 15 points and LatticeMind only 2.

\begin{table*}[t]
    \centering
    \small
    \caption{Software-state results. Top: overall scores on the two benchmarks. Bottom: per-metric breakdown on the harder budgeted long-noise benchmark (near-ceiling current-state retrieval is in the appendix). Higher is better.}
    \label{tab:software_state}
    \begin{tabular}{lNNNN}
        \toprule
        & {Concat} & {LLM-Merge} & {LLM-Merge-Inc} & {StateMemory} \\
        \midrule
        \multicolumn{5}{l}{\textbf{\textit{Overall scores}}} \\
        \addlinespace[0.3ex]
        Budgeted long-noise   & 0.52 & 0.51 & 0.70 & \cellbest{0.83} \\
        \rowtint Repo bridge (v2)      & 0.18 & \cellbest{0.61} & 0.41 & 0.51 \\
        \midrule
        \multicolumn{5}{l}{\textbf{\textit{Per-metric breakdown (budgeted long-noise)}}} \\
        \addlinespace[0.3ex]
        Scoped state          & 0.78 & 0.89 & 0.89 & \cellbest{0.89} \\
        \rowtint Provenance            & 0.50 & 0.00 & 0.61 & \cellbest{0.81} \\
        Temporal              & 0.23 & 0.18 & 0.38 & \cellbest{0.77} \\
        \rowtint Stale suppression     & 0.67 & 0.78 & 0.78 & \cellbest{0.78} \\
        \bottomrule
    \end{tabular}
\end{table*}

The per-metric breakdown in \cref{tab:software_state} (bottom) shows where the software-state gains come from after hardening the benchmark. The informative gap remains in provenance and temporal reasoning, which is where free-form summaries tend to compress away the evidence needed for auditable state tracking.

\subsection{Boundary on Coordination and Planning}
\label{subsec:boundary_public}

\begin{figure}[t]
    \centering
    \includegraphics[width=\columnwidth]{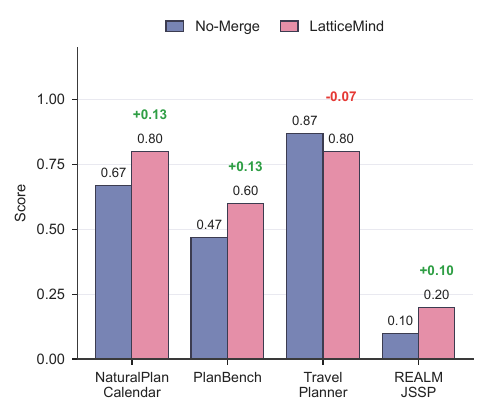}
    \caption{Secondary public benchmarks. Apples-to-apples comparison against \textsc{No-Merge}, the natural baseline that uses the same multi-agent pool but no reconciliation. LatticeMind matches or beats \textsc{No-Merge} on three of four tasks. The full table including \textsc{Single-Agent}, \textsc{Majority Vote}, \textsc{Debate}, and \textsc{Judge} is in \cref{tab:secondary_full}.}
    \label{fig:secondary_benchmarks}
\end{figure}

\Cref{fig:secondary_benchmarks} compares LatticeMind against \textsc{No-Merge}, the most apples-to-apples baseline since both methods use the same multi-agent pool and differ only in whether conflicts are reconciled. The reconciliation step adds 13 points on NaturalPlan Calendar (n=15) and 13 points on PlanBench (n=15), and shows a 10-point exploratory gain on REALM JSSP (n=10, 2/10 vs.\ 1/10), while it costs 7 points on TravelPlanner (n=10). The TravelPlanner regression is informative: the benchmark is a single-shot constraint problem that single-agent inference already solves at 0.97, so the multi-agent overhead introduces variance without conflicts to resolve. The REALM gap is small in absolute samples and we read it as exploratory rather than confirmatory.

The full per-method numbers in \cref{tab:secondary_full} show LatticeMind is top on PlanBench (0.60 vs.\ next-best 0.47) and third on NaturalPlan Calendar behind \textsc{Debate} and \textsc{Judge} (both 0.87); both deliberation methods exploit iterative refinement toward a globally consistent schedule, which one-shot reconciliation does not target.

\subsection{Application Study: Software-State Tracking}
\label{subsec:application_software}

\cref{tab:software_state} reports the application study under a harder query set. On the targeted budgeted long-noise benchmark, \textsc{StateMemory} reaches 0.83 overall, compared with 0.70 for the strongest summary baseline \textsc{LLM-Merge-Inc}. The gap is informative because all methods operate under the same document and reader budgets, and the queries stress provenance, temporal ordering, and stale suppression. The per-metric breakdown shows the gap is dominated by provenance, 0.81 against 0.61, and temporal reasoning, 0.77 against 0.38, which are the dimensions free-form summaries tend to compress away.

\begin{figure}[t]
    \centering
    \includegraphics[width=\columnwidth]{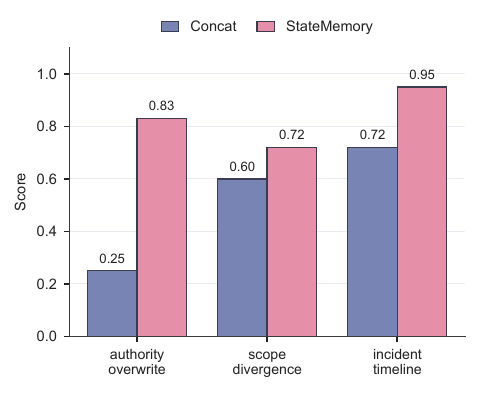}
    \caption{Per-family scores on the budgeted long-noise benchmark. \textsc{StateMemory} dominates \textsc{Concat} on every scenario family, with the largest absolute gap on \op{authority-overwrite} where stale claims must be suppressed.}
    \label{fig:per_family}
\end{figure}

\Cref{fig:per_family} confirms the pattern at the family level: \textsc{Concat} drops to 0.25 on \op{authority-overwrite} where stale claims must be actively suppressed, while \textsc{StateMemory} reaches 0.83. The harder repo bridge is challenging for the current instantiation (\textsc{LLM-Merge} 0.61 vs.\ \textsc{StateMemory} 0.51), suggesting the slot schema can miss sparse commit-level evidence a strong summarizer compresses well.

\subsection{Learned Operators on Unseen Repository Trajectories}
\label{subsec:learned_repo_ops}

\begin{figure}[t]
    \centering
    \includegraphics[width=\columnwidth]{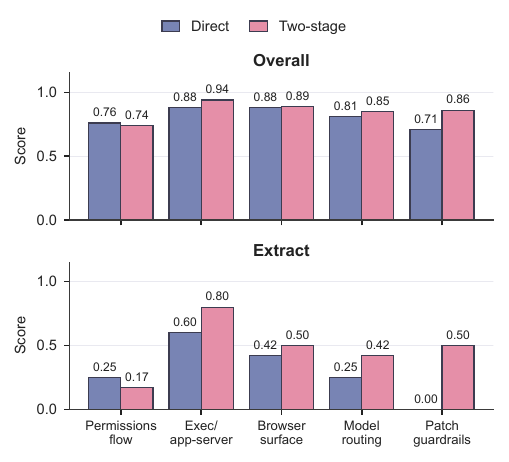}
    \caption{Five-way leave-one-out replay for learned operators. Top: overall; bottom: extraction-only. \op{compare-claims} and \op{update-memory} saturate at 1.00 in both settings and are omitted; full per-holdout breakdown in \cref{app:learned_ops}.}
    \label{fig:repo_operator}
\end{figure}

The Direct bars in \cref{fig:repo_operator} separate the operator-level story from the end-to-end bridge: \op{compare-claims} and \op{update-memory} saturate, so variation comes from overall transfer and especially \op{extract-state} (mean 0.30, worst case \op{patch-guardrails} at zero). The learned memory policy transfers; extraction does not yet match.

The Two-stage bars mitigate this: a constrained procedure that first selects supported entity-slot pairs and then generates values raises overall from 0.81 to 0.86 and extraction from 0.30 to 0.48, with the largest gain on \op{patch-guardrails} (0 to 0.50 extraction). Four of five holdouts improve; the single regression on \op{permissions-flow} shows constrained decoding can prune correct values when the direct extractor already grounds the right entity.

\subsection{Discussion}
\label{subsec:discussion}

LatticeMind wins when the bottleneck is deciding which incompatible claim to trust, and underperforms when it is search, deliberation, or holistic generation: ConflictBank is the cleanest case, and the ablation confirms the checker and reconciler are complementary. The secondary planning tasks map this boundary, where the memory-update advantage cannot fully fire yet LatticeMind still beats naive merge on three of four. On the software-state surface the gain concentrates in provenance and temporal questions; qualitative cases appear in \cref{app:qualitative}.

\section{Conclusion}\label{sec:conclusion}

We presented LatticeMind, a conflict-aware memory primitive that represents item status, evidence, and supersession, and combines cheap symbolic checks with selective LLM reconciliation. It substantially improves over single-agent and aggregation baselines on ConflictBank, and the same update rules transfer to long-horizon software-state tracking. A learned-operator study localizes the remaining gap to extraction rather than the update rule, pointing future work toward stronger extraction.

\section*{Limitations}

LatticeMind targets contradiction handling and is not a universal replacement for search or deliberation, as our secondary results illustrate. The evaluation uses a single hosted LLM with default decoding; the released runners record per-sample outcomes and draw their samples under a fixed seed, so any future model can be scored on the same examples in subsequent work.

\section*{Ethics Statement}

This work studies structured memory representations and conflict handling for LLM agents. It does not introduce a human-subject dataset, and the software-state bridge uses only public Codex repository metadata. By design, the rendered memory exposes provenance, contested states, and explicit status transitions to support human oversight; deployments in high-stakes settings should pair the rendered view with appropriate validation.

\bibliography{main}

\appendix
\section{Additional Experimental Details}
\label{app:details}

\subsection{Reproducibility Statement}
\label{app:reproducibility}

\paragraph{Artifacts.} The implementation comprises the LatticeMind memory, the benchmark runners, the prompt-based baselines, all prompt templates verbatim, and stand-alone analysis scripts that compute the bootstrap intervals and McNemar tests of \cref{app:stats} and draw \cref{fig:conflictbank,fig:secondary_benchmarks,fig:per_family}. Each runner records per-sample outcomes rather than aggregates alone, alongside the command, sample size, seed and model that produced them, so every reported statistic recomputes offline from a completed run without re-querying the hosted model. Two artifacts are not reusable in that way. The raw prompt/response logs embed ConflictBank evidence passages verbatim, and that corpus carries no redistribution license. The numbers behind \cref{fig:repo_operator} depend on QLoRA adapter checkpoints, so reproducing them requires re-running the fine-tuning rather than replaying a stored array.

\paragraph{Models and versions.} Prompt-based experiments use Qwen3-Max accessed through OpenRouter (snapshot of the public Qwen3-Max endpoint at submission time, default temperature). The learned-operator study uses 4-bit QLoRA adapters on \texttt{Qwen2.5-14B-Instruct}~\citep{qwen2024qwen25} with rank 16, alpha 32, applied to all attention and MLP projection matrices. Because hosted LLMs may evolve, exact reruns may vary with provider-side model updates unless the endpoint is version-pinned; the runners therefore record per-sample outcomes rather than aggregates alone, so every reported statistic can be recomputed from a run without re-querying the model for each figure.

\paragraph{Compute.} Prompt-based runs were executed on a single workstation, and per-sample cost is dominated by hosted-LLM API latency rather than local compute. The QLoRA adapter training used a single NVIDIA H100 80GB GPU. Each five-way leave-one-out run, including training and evaluation across all five folds, completes in roughly 6 to 8 GPU-hours, for an aggregate budget of approximately 40 GPU-hours.

\paragraph{Dataset and artifact licenses.} ConflictBank~\citep{su2024conflictbank} is released for research use by its authors. NaturalPlan~\citep{zheng2024naturalplan}, PlanBench~\citep{valmeekam2023planbench}, TravelPlanner~\citep{xie2024travelplanner}, and REALM-Bench~\citep{geng2025realmbench} are publicly released for academic evaluation under their respective project licenses. The Codex commit trajectories used in the repo-bridge are derived from a public open-source repository under its declared license; we replay only commit metadata (subject, date, touched files) and short hand-written summaries, never proprietary content. Unit tests cover the core lattice, merge, reconciler, belief, and software-state benchmark components.

\subsection{Public Benchmark Context}
\label{app:public_context}

Beyond ConflictBank, we evaluated LatticeMind on four public long-horizon tasks. \Cref{tab:secondary_full} reports every available method on each task; the main paper compares only against \textsc{No-Merge} for clarity. LatticeMind is the top method on PlanBench, beats \textsc{No-Merge} on three of four tasks, and is competitive on NaturalPlan Calendar where two stronger deliberation methods exploit the iterative-schedule structure of the task.

\begin{table}[h]
    \centering
    \small
    \caption{Full per-method scores on the four secondary public benchmarks. Calendar is exact match (n=15), PlanBench is goal-reached rate (n=15), TravelPlanner is constraint pass rate (n=10), REALM JSSP is validity rate (n=10). A dash means the baseline was not run on that task.}
    \label{tab:secondary_full}
    \begin{tabular}{lNNNN}
        \toprule
        Method & {Calendar} & {PlanBench} & {Travel} & {REALM} \\
        \midrule
        Single-Agent          & 0.20 & 0.27 & \cellbest{0.97} & \cellbest{0.20} \\
        \rowtint No-Merge              & 0.67 & 0.47 & 0.87 & 0.10 \\
        Majority Vote         & 0.13 & 0.40 & {---} & {---} \\
        \rowtint Debate                & \cellbest{0.87} & 0.40 & {---} & {---} \\
        Judge                 & \cellbest{0.87} & 0.20 & {---} & {---} \\
        \textbf{LatticeMind}  & 0.80 & \cellbest{0.60} & 0.80 & \cellbest{0.20} \\
        \bottomrule
    \end{tabular}
\end{table}

\subsection{Software-State Tracks}
\label{app:tracks}

The software-state benchmark includes four tracks:
\texttt{structured},
\op{raw-notes},
\texttt{budgeted}, and
\op{repo-bridge}.
The main paper emphasizes the \texttt{budgeted} track because it is the strongest stress test of memory under document constraints. In that track, the writer budget is 1700 characters and the reader budget is 1400 characters. The compact renderer is enabled so that all methods are compared under the same external memory budget rather than the internal size of their hidden data structures. In the latest harder version used in the paper, the query sets were expanded to include more provenance, rollback, scope-confusion, and stale-suppression questions, which lowers the earlier ceiling effects.

\begin{table}[h]
    \centering
    \footnotesize
    \caption{Full metric breakdown for the harder budgeted long-noise software-state benchmark. The current-state row is included here rather than in the main paper because it is already near ceiling for the stronger methods. SM = StateMemory.}
    \label{tab:software_breakdown_full}
    \begin{tabular}{lNNNN}
        \toprule
        Metric & {Concat} & {Merge} & {Merge-Inc} & {SM} \\
        \midrule
        Current state          & 0.78 & \cellbest{1.00} & \cellbest{1.00} & \cellbest{1.00} \\
        \rowtint Scoped state           & 0.78 & 0.89 & 0.89 & \cellbest{0.89} \\
        Provenance             & 0.50 & 0.00 & 0.61 & \cellbest{0.81} \\
        \rowtint Temporal               & 0.23 & 0.18 & 0.38 & \cellbest{0.77} \\
        Stale suppression      & 0.67 & 0.78 & 0.78 & \cellbest{0.78} \\
        \bottomrule
    \end{tabular}
\end{table}

\subsection{Scenario Families}
\label{app:scenario_families}

\paragraph{Authority overwrite.}
Sessions interleave code-backed changes with stale observations and outdated documentation. This family stresses whether the memory can suppress stale claims while preserving provenance to the winning update.

\paragraph{Scope divergence.}
Production, staging, and feature-preview states coexist. This family tests whether the memory preserves branch and environment scope rather than collapsing everything into a single project-wide summary.

\paragraph{Incident timeline.}
Rollback and restore events create non-monotonic state histories. This family emphasizes temporal questions such as when a rollback happened and which later commit restored the prior configuration.

\subsection{Real-Repository Bridge}
\label{app:bridge}

The bridge experiment replays two real Codex commit trajectories:
(i) a permissions-flow rollout and
(ii) an exec/app-server rollout.
Each replay session contains the real commit date, subject, a short summary, touched-file hints, and a structured state card derived from the commit. We additionally interleave later stale notes that make plausible but wrong inferences about the rollout, so the bridge now probes whether a memory system can retain the authoritative change history while suppressing late contradictory observations. In the harder v2 used by the final paper draft, each case receives extra distractor sessions and more multi-hop provenance/temporal questions; under this setting, \textsc{LLM-Merge} reaches 0.61 overall, \textsc{StateMemory} 0.51, \textsc{LLM-Merge-Inc} 0.41, and \textsc{Concat} 0.18. The bridge is therefore more realistic than purely synthetic scenarios, but it is still a curated replay rather than an unrestricted benchmark over raw diffs.

\subsection{Learned Operator Study}
\label{app:learned_ops}

To separate the memory update rule from the extraction bottleneck, we trained QLoRA adapters on \texttt{Qwen2.5-14B-Instruct} for three software-state operators:
\op{extract-state},
\op{compare-claims}, and
\op{update-memory}.
Training data mixes the synthetic software-state scenarios with replay-derived repo examples, and evaluation uses five leave-one-out Codex trajectories:
\op{permissions-flow},
\op{exec-app-server},
\op{browser-surface},
\op{model-routing}, and
\op{patch-guardrails}.
Under schema-only normalization, the learned adapters are already at ceiling on the current \op{compare-claims} and \op{update-memory} evaluation across all five holdouts, while \op{extract-state} averages 0.30 claim-F1. We therefore interpret the learned study as evidence that the conflict/update abstraction itself transfers, while extraction remains the limiting factor.

\begin{table}[h]
    \centering
    \small
    \caption{Full per-holdout learned-operator breakdown for the five-way leave-one-out repository replay study. The compare/update columns saturate to 1.0 in both direct and two-stage settings; the main paper folds these into \cref{fig:repo_operator}.}
    \label{tab:repo_operator_loo_full}
    \begin{tabular}{lNNNN}
        \toprule
        Holdout & {Overall} & {Extract} & {Compare} & {Update} \\
        \midrule
        Permissions flow         & 0.76 & 0.25 & 1.00 & 1.00 \\
        \rowtint Exec/app-server          & \cellbest{0.88} & \cellbest{0.60} & 1.00 & 1.00 \\
        Browser surface          & 0.88 & 0.42 & 1.00 & 1.00 \\
        \rowtint Model routing            & 0.81 & 0.25 & 1.00 & 1.00 \\
        Patch guardrails         & 0.71 & 0.00 & 1.00 & 1.00 \\
        \midrule
        \textbf{Mean}            & \bfseries 0.81 & \bfseries 0.30 & \bfseries 1.00 & \bfseries 1.00 \\
        \bottomrule
    \end{tabular}
\end{table}

The constrained two-stage decoder helps most when the direct extractor emits the wrong entity family or collapses under longer evidence strings. This is why \op{patch-guardrails} gains the most: the direct extractor often drifts into neighboring approval or TUI entities, while the two-stage extractor at least selects the correct \op{patch-runtime} slots before attempting value generation.

\begin{figure*}[h]
    \centering
    \includegraphics[width=0.92\textwidth]{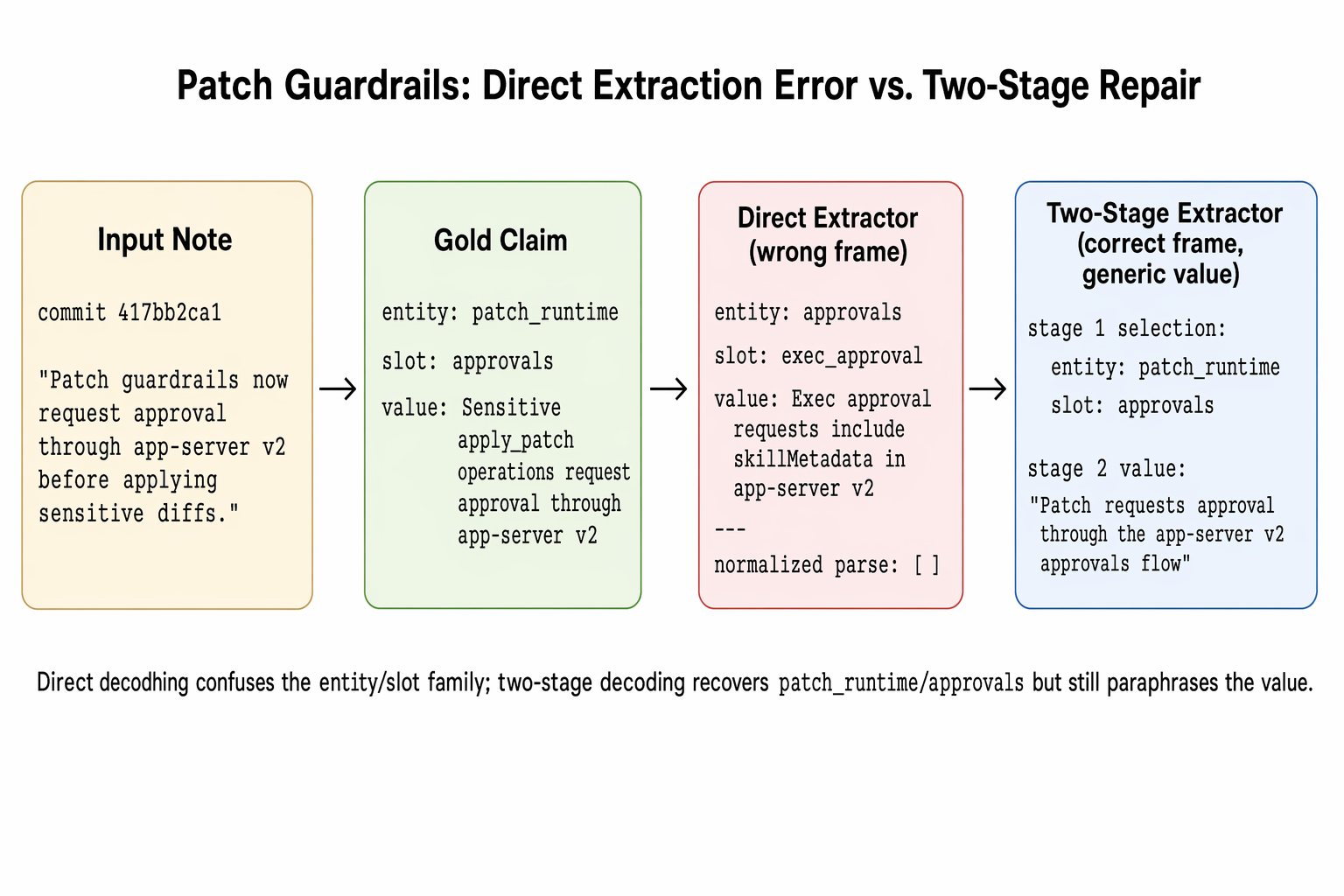}
    \caption{Representative \op{patch-guardrails} extraction example. The direct extractor drifts into the wrong entity/slot family and normalizes to an empty claim set, while the two-stage extractor recovers the correct \texttt{patch\_runtime/approvals} frame and only misses on value specificity.}
    \label{fig:patch_two_stage_case}
\end{figure*}

\paragraph{Representative error pattern.}
On \op{patch-guardrails}, the direct extractor often maps the note about guarded \texttt{apply\_patch} execution onto the wrong entity family, producing malformed or irrelevant \texttt{exec}/\texttt{tui} claims and yielding zero extraction credit. The two-stage extractor fixes the entity-slot selection and recovers \texttt{patch\_runtime.execution}, but it can still generate an overly generic value such as ``patch requests use the experimental tool-call approval flow'' instead of the gold ``\texttt{apply\_patch} runs through the structured diff engine.'' On \op{browser-surface}, by contrast, the two-stage extractor often recovers both the \texttt{browser\_runtime.execution} and \texttt{browser\_runtime.file\_storage} slots from the same note, which is enough to move the example from partial credit to full semantic match. These cases reinforce the same conclusion as the aggregate numbers: extraction errors are now mostly about grounding the right value, not about conflict/update logic. \Cref{fig:patch_two_stage_case} visualizes a representative \op{patch-guardrails} example.

\subsection{Earlier Open-Ended Coding-Memory Iteration}
\label{app:earlier_iteration}

Before building the current software-state benchmark, we also ran an earlier open-ended coding-memory benchmark over multi-session project evolution scenarios. That earlier setup often favored free-form summarization baselines and did not consistently show an advantage for the structured belief pipeline. In hindsight, it rewarded broad overview generation more than precise stale-suppression, scope preservation, or provenance retrieval. This negative result directly motivated the narrower software-state benchmark used in the main paper.

\subsection{Qualitative Case Studies}
\label{app:qualitative}

\Cref{tab:qualitative_cases} shows two replayed cases from the real Codex bridge. We use real commit trajectories rather than synthetic scenarios because the qualitative value of LatticeMind is easiest to see when several small implementation changes must be turned into a stable operational picture. In both cases, the rendered memory is compact enough to fit within the benchmark budget while still preserving the commit-level provenance that downstream questions require.

\begin{table*}[t]
    \centering
    \footnotesize
    \caption{Qualitative case studies from replayed real repository trajectories. Each row shows the same progression: input sessions from real commits, the compact rendered memory produced by \textsc{StateMemory}, and representative downstream outputs.}
    \label{tab:qualitative_cases}
    \renewcommand{\arraystretch}{1.15}
    \begin{tabular}{>{\raggedright\arraybackslash}p{0.27\textwidth}>{\raggedright\arraybackslash}p{0.37\textwidth}>{\raggedright\arraybackslash}p{0.28\textwidth}}
        \toprule
        Input sessions & Rendered memory state & Example output \\
        \midrule
        \textbf{Case 1: permissions flow rollout} \newline
        \texttt{e6b93841c}: add built-in \texttt{request\_permissions} tool and forward approvals through app-server v2 \newline
        \texttt{340f9c9ec}: add \texttt{skillMetadata} to exec approval requests \newline
        \texttt{06f82c123}: TUI renders the approval overlay
        &
        \textbf{Current state} \newline
        \texttt{permissions\_flow.approvals} = built-in \texttt{request\_permissions} via \texttt{requestApproval} \newline
        \texttt{permissions\_flow.ui\_integration} = TUI renders approval overlay \newline
        \texttt{exec\_approval.approvals} = exec requests include \texttt{skillMetadata} \newline
        \textbf{Transitions} \newline
        tool introduced in \texttt{e6b93841c}; UI surface added later in \texttt{06f82c123}
        &
        \textbf{Query answer} \newline
        ``Which commit introduced the built-in tool?'' \(\rightarrow\) \texttt{e6b93841c} \newline
        \textbf{Task output} \newline
        \op{permissions-flow}: running turns use \texttt{request\_permissions} and forward approval through app-server v2 \newline
        \op{ui-surface}: TUI renders the approval overlay
        \\
        \midrule
        \textbf{Case 2: exec/app-server rollout} \newline
        \texttt{a684a3609}: app-server hot-reloads user config after batch write \newline
        \texttt{da3689f0e}: non-interactive exec switches to an in-process app server \newline
        \texttt{340f9c9ec}: exec approvals carry \texttt{skillMetadata}
        &
        \textbf{Current state} \newline
        \texttt{exec\_runtime.app\_server} = in-process app server \newline
        \texttt{exec\_runtime.execution} = non-interactive exec uses app server \newline
        \texttt{app\_server.config\_reload} = hot-reload user config \newline
        \texttt{exec\_approval.approvals} = include \texttt{skillMetadata} \newline
        \textbf{Transitions} \newline
        config hot-reload appears before the exec runtime switch
        &
        \textbf{Query answer} \newline
        ``What behavior was added before exec switched?'' \(\rightarrow\) hot-reload user config, commit \texttt{a684a3609} \newline
        \textbf{Task output} \newline
        \texttt{runtime}: exec uses the in-process app server \newline
        \texttt{approvals}: exec approval requests carry \texttt{skillMetadata} \newline
        \texttt{config}: app-server hot-reloads user config
        \\
        \bottomrule
    \end{tabular}
\end{table*}

These cases illustrate a practical difference between structured state memory and summary-only approaches. The summary baselines can often restate the current architecture in fluent prose, but they do not reliably keep the exact commit attribution aligned with the correct change boundary. The rendered LatticeMind state, by contrast, makes the causal chain explicit enough that a downstream reader can answer both ``what is true now?'' and ``which change introduced it?'' from the same compact document.

\subsection{Implementation Details}
\label{app:implementation}

\paragraph{Memory item state machine.}
Each memory item carries a status \(\eta \in \{\textsc{Proposed}, \textsc{Confirmed}, \textsc{Contested}, \textsc{Superseded}\}\). All items start as \textsc{Proposed}; the symbolic checker can transition to \textsc{Contested}; the reconciler transitions to \textsc{Confirmed} or \textsc{Superseded}; and \textsc{Superseded} is terminal (backward transitions are forbidden). State transitions are applied centrally by the orchestrator so that the rendered view reflects only confirmed and explicitly contested items.

\paragraph{Symbolic checker.}
The checker performs two cheap structural passes. First, it builds a directed graph from \texttt{DEPENDENCY} edges and rejects any cycle (NetworkX strongly-connected-component scan). Second, it parses \texttt{CONSTRAINT} content with the regex \texttt{resource:(\textbackslash{}S+)\textbackslash{}s+time:(\textbackslash{}d+)-(\textbackslash{}d+)} and reports an interval overlap whenever \(s_1 < e_2\) and \(s_2 < e_1\) for two intervals over the same resource (so adjacent boundaries do not collide). Each detected violation is emitted with the IDs of the two conflicting entries; the reconciler then decides whether to supersede or to retain both as a coordination signal.

\paragraph{Evidence weight table.}
\Cref{tab:evidence_weights} lists the full evidence-type weights used in the software-state score \(s(z) = w(\op{evidence-type}) + 40 \cdot \mathbf{1}[\texttt{git\_commit} \neq \emptyset]\). Strong code-backed signals outrank notes and especially explicitly stale observations. The git-backed bonus dominates ordinary type differences, which lets a freshly committed change supersede a stale note without further LLM judgment.

\begin{table}[h]
    \centering
    \small
    \caption{Evidence-type weights used by the software-state resolver. Higher means stronger evidence.}
    \label{tab:evidence_weights}
    \begin{tabular}{lS[table-format=2.0]}
        \toprule
        Evidence type & {Weight} \\
        \midrule
        \op{code-change}              & \cellbest{60} \\
        \rowtint \op{incident-hotfix}          & \cellbest{60} \\
        \op{config-observation}       & 45 \\
        \rowtint \op{runtime-observation}      & 40 \\
        \op{branch-experiment}        & 25 \\
        \rowtint \op{human-note}               & 12 \\
        \op{stale-observation}        & 4 \\
        \midrule
        Git-commit bonus \(\mathbf{1}[\texttt{git\_commit} \neq \emptyset]\) & {$+40$} \\
        \bottomrule
    \end{tabular}
\end{table}

\paragraph{ConflictBank agents and prompts.}
Each ConflictBank example is processed by three agents, each reading exactly one of the three sources. All three agents share an identical extraction prompt asking for structured claims and evidence-type tags without choosing a final answer; per-agent identifiers used in the implementation are bookkeeping labels and are never exposed to the agent itself, to the reader, or to the reconciler. The reconciler uses the adaptive single-pass prompt that first classifies a conflict as \textsc{Credibility} or \textsc{Coordination} and supersedes losers only in the \textsc{Credibility} branch. Crucially, neither the extractor nor the reconciler receives any role label or source-name hint indicating which source is the authoritative, alternative, or recent one. Source order is preserved across examples for replay determinism, but the reconciler's policy is stateless and content-driven: it scores a claim from the evidence-type tags inferred from each source's text (e.g., explicit citations, recency markers, or self-described source kind), not from per-example position. The label-blind diff in \cref{tab:label_blind_diff} also serves as an indirect robustness check: removing the role names from the reader prompt costs every baseline 8 to 15 accuracy points while costing LatticeMind only 2, which is the opposite of what would happen if LatticeMind relied on a positional or role prior.

\paragraph{Same-pool baseline parity.}
All ConflictBank methods see the same three sources presented under the same neutral identifiers. \textsc{Single-Agent} reads all three in one prompt; \textsc{No-Merge}, \textsc{Majority Vote}, \textsc{Debate}, and \textsc{Judge} all consume the same three per-source extractions and differ only in how answers are aggregated. The structural overhead added by LatticeMind is the symbolic checker (negligible cost) and the LLM reconciler, invoked an average of 1.29 times per example on the 75-sample run. The comparison is therefore not between richer prompting and weaker prompting but between explicit persistent conflict resolution and answer-only aggregation over the same evidence pool. The evidence-label question is isolated by \cref{tab:label_blind_diff} rather than by the ablation table: removing role names from the reader prompt costs every answer-level baseline 8 to 15 points while costing LatticeMind 2, so the gain is not the evidence-tag heuristic itself but the act of resolving incompatible claims at write time. The final row of \cref{tab:conflictbank_ablation} is a repeat of the full configuration, not an evidence ablation; see \cref{subsec:ablation}.

\paragraph{Reconciler invocation rate.}
On the 75-sample ConflictBank run, the symbolic checker reports an average of 1.29 violations per example after per-resource grouping (multiple overlapping constraints on the same resource are summarized into one violation), and the reconciler is invoked once per remaining violation. On the 50-sample ablation, the full configuration, its repeat, and $-$reconciler each report 3.0 ungrouped pairwise violations per example (one per pair among three agents). The $-$checker ablation reports 0 violations and so triggers no reconciliation, while $-$reconciler surfaces violations but cannot resolve them, raising the post-hoc conflict rate to 0.08 or 0.10.

\subsection{Label-blind vs.\ Label-aware ConflictBank}
\label{app:label_blind_diff}

\Cref{tab:label_blind_diff} reports the impact of the label-blind protocol on every method. Renaming the three sources to neutral identifiers (Source A/B/C) and removing the trust hint from the reader prompt costs every baseline 8 to 15 accuracy points. LatticeMind loses only 2 points, which is consistent with our claim that LatticeMind reads the resolved memory status rather than the source name. The gap between LatticeMind and the strongest answer-level baseline widens from 23 to 36 points under the label-blind protocol.

\begin{table}[h]
    \centering
    \small
    \caption{Effect of the label-blind protocol on ConflictBank accuracy (75 samples). Original = source names visible and trust hint present in reader prompt. Label-blind = neutral Source A/B/C names, no trust hint. LatticeMind is robust to the change; every baseline drops substantially.}
    \label{tab:label_blind_diff}
    \begin{tabular}{lNNS[table-format=-1.2]}
        \toprule
        Method & {Original} & {Label-blind} & {$\Delta$} \\
        \midrule
        Single-Agent          & 0.72 & 0.63 & -0.09 \\
        \rowtint No-Merge              & 0.44 & 0.32 & -0.12 \\
        Majority Vote         & 0.76 & 0.61 & -0.15 \\
        \rowtint Debate                & 0.57 & 0.45 & -0.12 \\
        Judge                 & 0.68 & 0.60 & -0.08 \\
        \textbf{LatticeMind}  & \cellbest{0.99} & \cellbest{0.97} & \cellbest{-0.02} \\
        \bottomrule
    \end{tabular}
\end{table}

\subsection{Statistical Analysis}
\label{app:stats}

We complement the headline accuracies with bootstrap confidence intervals and paired McNemar tests on per-sample correctness. Bootstrap CIs use 10{,}000 resamples with a fixed seed over the per-example correctness arrays written by the benchmark runners; McNemar uses the exact two-sided binomial on the discordant pair counts \((n_{01}, n_{10})\). \Cref{tab:stats_main} reports the main label-blind ConflictBank comparison and \cref{tab:stats_ablation} reports the single-pass ablation.

\begin{table}[h]
    \centering
    \small
    \caption{ConflictBank bootstrap 95\% CI and McNemar tests vs.\ LatticeMind. Discordant pairs $(n_{01}, n_{10})$ count examples where the baseline is wrong/right when LatticeMind is right/wrong.}
    \label{tab:stats_main}
    \begin{tabular}{lcc}
        \toprule
        Method & Acc.\ [95\% CI] & McNemar $p$ ($n_{01}/n_{10}$) \\
        \midrule
        Single-Agent          & 0.63 [0.52, 0.73] & $2.2{\cdot}10^{-7}$\,\,(27/1) \\
        \rowtint No-Merge              & 0.32 [0.21, 0.43] & $3.6{\cdot}10^{-15}$\,\,(49/0) \\
        Majority Vote         & 0.61 [0.51, 0.72] & $1.1{\cdot}10^{-7}$\,\,(28/1) \\
        \rowtint Debate                & 0.45 [0.35, 0.57] & $3.8{\cdot}10^{-11}$\,\,(40/1) \\
        Judge                 & 0.60 [0.49, 0.71] & $5.8{\cdot}10^{-8}$\,\,(29/1) \\
        \rowbest \textbf{LatticeMind}  & \textbf{0.97 [0.93, 1.00]} & \textit{reference} \\
        \bottomrule
    \end{tabular}
\end{table}

Every aggregation baseline is rejected at $p<10^{-6}$ (worst case $2.2\cdot10^{-7}$ for Single-Agent), with discordant counts that are highly asymmetric: e.g.\ \textsc{No-Merge} is wrong on 49 examples that LatticeMind gets right and never wins on cases LatticeMind misses. The narrow CI around 0.97 reflects 73/75 correct rather than a small-sample artifact; even under the most adversarial bootstrap resamples the lower bound stays at 0.93.

\begin{table}[h]
    \centering
    \small
    \caption{Single-pass ablation, 50 samples. Bootstrap 95\% CI and McNemar tests vs.\ \textit{full}.}
    \label{tab:stats_ablation}
    \begin{tabular}{lcc}
        \toprule
        Configuration & Acc.\ [95\% CI] & McNemar $p$ ($n_{01}/n_{10}$) \\
        \midrule
        \rowbest \textbf{Full}         & \textbf{0.96 [0.90, 1.00]} & \textit{reference} \\
        $-$checker            & 0.84 [0.74, 0.94] & 0.07\,\,(7/1) \\
        \rowtint $-$reconciler         & 0.82 [0.70, 0.92] & 0.04\,\,(8/1) \\
        Full (repeat)         & 0.96 [0.90, 1.00] & 1.00\,\,(0/0) \\
        \bottomrule
    \end{tabular}
\end{table}

The reconciler ablation is significant at $p<0.05$ and the checker ablation is borderline at $p=0.07$. Both directions are consistent with the headline 12 to 14 point drops, but the 50-sample size limits power on individual flips. The repeat of the full configuration produces zero discordant pairs, which bounds run-to-run variance on this evidence pool; it is a stability check, not an ablation.

\paragraph{Caveats.}
The reported $p$-values are conditioned on a single decoding run with Qwen3-Max. Because each runner records per-example outcomes, the McNemar and bootstrap tests in \cref{tab:stats_main,tab:stats_ablation} recompute deterministically from a run rather than being reported as bare aggregates. The discordant-pair counts are highly asymmetric (e.g.\ 49/0 against \textsc{No-Merge}), so on this evidence pool the comparative conclusions are robust to per-example flips. Generalization across model families and decoding seeds is a deliberate next step rather than a contradicted claim; the released runners score any future model or baseline on the same sampled examples under a fixed seed.

\paragraph{Models, infrastructure, and tooling.}
The prompt-based memory benchmarks reported in the main paper use Qwen3-Max for both memory operations and reader/task evaluation; calls go through OpenRouter with default temperature and a fixed system message per role. The learned-operator study uses local \texttt{Qwen2.5-14B-Instruct} QLoRA adapters (rank 16, alpha 32, all attention and MLP projections, 4-bit quantization). Statistical tests are recomputed from the per-sample JSONs a run produces, by a stand-alone script in the repository (no hidden random state beyond a fixed bootstrap seed). Unit and benchmark tests pass in the current repository snapshot.

\end{document}